\documentclass[preprint,12pt, preprint]{elsarticle}

\usepackage[T1]{fontenc}
\usepackage{multirow}
\usepackage{multicol}
\usepackage{graphicx}
\usepackage{amssymb}
\usepackage{booktabs}
\usepackage[normalem]{ulem}
\useunder{\uline}{\ul}{}
\usepackage{url}

\journal{AI Open}

\begin{document}

\begin{frontmatter}



\title{Large Language Models for Missing Data Imputation: Understanding Behavior, Hallucination Effects, and Control Mechanisms}

\author[1,2]{Arthur Dantas Mangussi\corref{cor1}}
% \cormark[1] % Corresponding author indication
\cortext[cor1]{Corresponding author}
\ead{arthuradm@ita.br}

\author[3]{Ricardo Cardoso Pereira}
\ead{rdpereira@dei.uc.pt}

\author[1,2]{Ana Carolina Lorena}
\ead{aclorena@ita.br}

\author[3]{Pedro Henriques Abreu}
\ead{pha@dei.uc.pt}

\affiliation[1]{organization={Computer Science Division, Aeronautics Institute of Technology},
            addressline={Praça Marechal Eduardo Gomes, 50}, 
            city={São José dos Campos},
            postcode={12228-900}, 
            state={São Paulo},
            country={Brazil}}
            
\affiliation[2]{organization={Science and Technology Institute, Federal University of São Paulo},
            addressline={Avenue Cesare Monsueto Giulio Lattes, 1201}, 
            city={São José dos Campos},
            postcode={12231-280}, 
            state={São Paulo},
            country={Brazil}}

\affiliation[3]{organization = {University of Coimbra, CISUC/LASI – Centre for Informatics and Systems of the University of Coimbra, Department of Informatics Engineering},
        addressline = {Polo II, Pinhal de Marrocos},
        city = {Coimbra},
        citysep={},
        postcode={3030-290},
        country={Portugal}}

%% Abstract
\begin{abstract}
Data imputation is a cornerstone technique for handling missing values in real-world datasets, which are often plagued by missingness. Despite recent progress, prior studies on Large Language Models-based imputation remain limited by scalability challenges, restricted cross-model comparisons, and evaluations conducted on small or domain-specific datasets. Furthermore, heterogeneous experimental protocols and inconsistent treatment of missingness mechanisms (MCAR, MAR, and MNAR) hinder systematic benchmarking across methods. This work investigates the robustness of Large Language Models for missing data imputation in tabular datasets using a zero-shot prompt engineering approach. To this end, we present a comprehensive benchmarking study comparing five widely used LLMs against six state-of-the-art imputation baselines. The experimental design evaluates these methods across 29 datasets (including nine synthetic datasets) under Missing Completely at Random, Missing at Random, and Missing Not at Random mechanisms, with missing rates of up to 20\%. The results demonstrate that leading LLMs, particularly Gemini 3.0 Flash and Claude 4.5 Sonnet, consistently achieve superior performance on real-world open-source datasets compared to traditional methods. However, this advantage appears to be closely tied to the models' prior exposure to domain-specific patterns learned during pre-training on internet-scale corpora. In contrast, on synthetic datasets, traditional methods such as MICE outperform LLMs, suggesting that LLM effectiveness is driven by semantic context rather than purely statistical reconstruction. Furthermore, we identify a clear trade-off: while LLMs excel in imputation quality, they incur significantly higher computational time and monetary costs. Overall, this study provides a large-scale comparative analysis, positioning LLMs as promising semantics-driven imputers for complex tabular data.
\end{abstract}


% \begin{highlights}
% \end{highlights}

%% Keywords

\end{frontmatter}



\section{Introduction} \label{sec:intro}

Data mining is defined as the process of uncovering hidden patterns in data, extracting valuable information, and transforming it into actionable knowledge~\cite{Witten2005}. In real-world domains, data typically presents several inconsistencies, such as class imbalance, class overlap, noise, and missing data, among others~\cite{Clemente2023}. Missing Data (MD) is characterized by the absence of information in one or more features within a dataset. The literature categorizes missing data into three fundamental mechanisms: Missing Completely at Random (MCAR), Missing at Random (MAR), and Missing Not at Random (MNAR)~\cite{Rubin1976}. These mechanisms can be described as follows~\cite{mdatagen2025}:

\begin{itemize}
    \item \textbf{MCAR}: the data missingness process occurs randomly without any dependency on specific features within the dataset;
    \item \textbf{MAR}: a dependency between existing features determines the missingness nature;
    \item \textbf{MNAR}: the missing values are related to themselves or to unobserved external data.
\end{itemize}

Most Machine Learning (ML) classifiers do not explicitly address the missing data problem during the training phase. As a result, the literature proposes several strategies to handle MD, with data imputation being the most widely adopted approach~\cite{mdatagen2025}. Missing Data Imputation refers to the process of replacing missing entries with estimated values according to a predefined strategy~\cite{Mangussi2024Adv}. The simplest imputation techniques rely on descriptive statistical measures, such as mean or median imputation. In contrast, more advanced and robust approaches leverage Deep Learning (DL)-based methods, including autoencoders (AEs) and Generative Adversarial Networks (GANs), to model complex data distributions and dependencies~\cite{Hasan2021,Sun2023Deep}.

Recently, Artificial Intelligence (AI) techniques have demonstrated the capability to generate seemingly novel and meaningful content (such as text, images, or audio) by learning patterns from large-scale training data. These advances have given rise to a research field commonly referred to as Generative Artificial Intelligence~\cite{Feuerriegel2023}. Within this context, the literature has increasingly explored a class of neural networks trained on massive volumes of data, known as Large Language Models (LLMs). LLMs are typically built upon the transformer architecture and rely on attention mechanisms, being pre-trained via self-supervised learning objectives in which auxiliary tasks, such as next-token prediction, are designed to learn rich representations of natural language while mitigating overfitting. This pre-training phase leverages internet-scale textual corpora, enabling LLMs to acquire extensive semantic and syntactic knowledge. As a result, LLMs have been successfully employed across a wide range of research problems in natural language processing~\cite{Feuerriegel2023}.

Within the MD domain, LLMs have increasingly been adopted for data imputation tasks. Subham Jha, Aman Goyal, and Rakesh Meena~\cite{Subham2026} provide a comprehensive survey of LLM-based approaches for missing data imputation. As highlighted in their survey and in prior literature (discussed in greater detail in Section~\ref{sec:related_works}) most existing methods reformulate tabular data into natural language representations to leverage the reasoning capabilities of LLMs. Despite recent progress, prior studies on LLM-based imputation remain limited by scalability challenges, restricted cross-model comparisons, and evaluations conducted on small or domain-specific datasets. Furthermore, heterogeneous experimental protocols and inconsistent treatment of missingness mechanisms (MCAR, MAR, and MNAR) hinder systematic benchmarking across methods. These limitations highlight important gaps in the literature and motivate the following research questions:

\begin{itemize}
\item \textbf{RQ1}: Are LLMs capable of performing robust missing data imputation through prompt engineering alone or can it bias the results somehow?
\item \textbf{RQ2}: Does the extensive knowledge encoded in internet-scale corpora of LLMs add value to missing data imputation tasks?
\item \textbf{RQ3}: Are hallucinations in LLMs more likely to occur in imputation contexts that are unfamiliar to the model?
\end{itemize}

Based on these research questions, this work aims to investigate the robustness of different LLMs for missing data imputation tasks. Our hypothesis is that, since LLMs are large-scale neural networks trained on vast internet corpora, many widely used open-source datasets, such as those available in the UCI Machine Learning Repository\footnote{\url{https://archive.ics.uci.edu/}}, are likely to be implicitly represented within their learned knowledge. Consequently, we hypothesize that LLMs can leverage this prior knowledge to outperform state-of-the-art imputation methods under comparable experimental settings. 

To reach these findings, we developed an experimental setup using eleven imputation methods, encompassing five frequently used LLMs models, under MCAR, MAR, and MNAR mechanisms up to 5\%, 10\%, and 20\% missing rates. Additionally, our work has investigated a foundation model for missing data imputation. For the evaluation, we used the Normalized Root Mean Square Error (NRMSE). Our results show that LLMs exhibit overall optimistic performance when used for tabular data imputation in open-source real-world datasets, likely due to the extensive background knowledge encoded during large-scale pretraining. In contrast, these models do not achieve the same level of performance in private or domain-specific scenarios, suggesting that the information available in large-scale internet corpora plays an important role in enabling LLMs to infer missing values. Furthermore, when the contextual information is unfamiliar or insufficient, LLMs tend to produce hallucinations, which can negatively affect the reliability of the imputed data. The main findings of this work are:
\begin{itemize}
    \item A prompt engineering approach that leverages the background knowledge encoded in LLMs for open-source datasets, which can outperform traditional data imputation methods;
    \item Prior knowledge derived from large-scale internet corpora has a significant impact on the effectiveness of LLM-based data imputation;
    \item Semantic context plays a crucial role in LLM-based imputation, as these models rely on familiar contextual information to improve the quality of the imputed values.
\end{itemize}

The remainder of this work is organized as follows: Section \ref{sec:related_works} presents the prior literature on missing data imputation using LLMs; Section \ref{sec:methodology} outlines the methodology applied in this study; Section \ref{sec:results} presents the results and discussion; and Section \ref{sec:conclusion} concludes the paper.

\section{Related Works} \label{sec:related_works}

Recent research investigating the use of LLMs for Missing Data (MD) imputation can be organized into four main methodological paradigms: (i) prompt-based direct imputation, (ii) hybrid architectures integrating LLMs with structural modules, (iii) context-enrichment frameworks leveraging LLM-generated semantic descriptors, and (iv) transformer-based models specifically designed for tabular data. This taxonomy provides a clearer conceptual organization of the field and highlights the methodological differences underlying existing approaches.

\subsection{Prompt-Based Direct Imputation}

The first line of work employs LLMs as direct generative predictors of missing values through prompt engineering. In this paradigm, tabular records are transformed into natural language prompts, and the LLM predicts the missing attribute in a zero-shot or fine-tuned setting.

Ding et al.~\cite{Ding2024} fine-tune a distilled GPT-2 model using a Low-Rank Adaptation (LoRA) strategy to perform missing value prediction in recommendation systems. The imputation task is reformulated as a question-answering problem, where structured tabular entries are converted into textual prompts. Their evaluation on AdClick and MovieLens datasets demonstrates improvements over classical imputation baselines such as Mean, $k$NN, and Multivariate Imputation. However, the approach requires iterative querying for each missing entry, raising concerns regarding scalability for large tabular datasets.

Similarly, Nazir, Cheema, and Wang~\cite{Nazir2023} use ChatGPT as a direct imputation engine on the Human Connectome Project dataset. By embedding domain-specific context within prompts, the model predicts missing features individually. While ChatGPT outperforms traditional statistical methods in their experiments, the evaluation is limited to a single dataset, and the one-value-at-a-time imputation procedure may hinder computational efficiency at scale.

These approaches share common characteristics: (i) reliance on natural language reformulation of structured data, (ii) iterative querying per missing value, and (iii) limited evaluation under standardized experimental protocols across diverse missingness mechanisms.

\subsection{Hybrid Architectures Integrating LLMs with Structural Modules}

A second research direction integrates LLMs into broader neural architectures designed to capture structural dependencies inherent to specific data domains.

Chen and Xu~\cite{Chen2023} propose Graph Attention Network Generative Pre-trained Transformer (GATGPT), which combines a pre-trained GPT-2 model with a graph attention mechanism for spatiotemporal imputation. By embedding graph-based representations into the language modeling process, the framework enhances the modeling of spatial dependencies. Experimental results across three benchmark datasets show competitive performance relative to statistical, machine learning, and deep learning baselines.

Le et al.~\cite{LeFang2025} introduce a spatiotemporal pre-trained LLM (SPLLM) architecture tailored for forecasting under missing data conditions. The model incorporates a feed-forward fine-tuning strategy, a spatiotemporal fusion graph convolutional module, and a final fusion layer. Their evaluation under MCAR scenarios with varying missing rates demonstrates consistent improvements over multiple baselines.

While hybrid approaches effectively model domain-specific structures, they are typically designed for particular data modalities (e.g., sensor networks or spatiotemporal forecasting) and therefore do not constitute general-purpose frameworks for tabular imputation.

\subsection{Context-Enrichment and Dual-Phase Frameworks}

A third paradigm leverages LLMs to generate semantic descriptors or contextual enrichments prior to downstream imputation. Instead of directly predicting numerical values, these approaches first construct contextually enhanced representations of missing attributes.

Hayat and Hasan~\cite{Hayat2024} propose the Contextual Language model for Accurate Imputation Method (CLAIM), a dual-phase framework in which LLMs generate natural language descriptions for missing entries. These enriched representations are subsequently used to fine-tune models for downstream tasks. Their experiments consider MCAR, MAR, and MNAR mechanisms and demonstrate improved performance over traditional methods such as Mean imputation, $k$NN, MICE, missForest, and GAIN. However, the method assumes that the generated descriptors sufficiently capture the underlying data semantics, and the evaluation does not address large-scale datasets.

Extending this idea, the same authors introduce Contextually Relevant Imputation leveraging pre-trained Language Models (CRILM)~\cite{Hayat2025}. CRILM follows a similar dual-phase process, where LLM-generated descriptors are used to construct enriched datasets, followed by missingness-aware fine-tuning of a smaller language model. Evaluations across six UCI datasets under MCAR, MAR, and MNAR scenarios show competitive and stable performance.

Although context-enrichment frameworks mitigate the need for direct value prediction, they remain dependent on prompt design and semantic interpretation, and their scalability to large tabular datasets remains insufficiently explored.

\subsection{Transformer-Based Models for Tabular Imputation}

In parallel to general-purpose LLM approaches, several works investigate transformer architectures specifically designed or adapted for tabular data imputation.

Murthy and Kumar~\cite{Murthy2025} conduct a systematic comparison of imputation strategies, including a BERT-like transformer model adapted for tabular inputs. Their findings suggest that transformer-based models can achieve strong predictive performance relative to classical methods.

Wei et al.~\cite{Wei2024} propose Table Transformers for Imputing Textual Attributes (TTITA), a transformer architecture designed to impute unstructured textual columns using structured tabular variables. Their experiments compare TTITA against Mode imputation, Mistral, Llama2, ChatGPT, $k$NN, LSTM, Transformer Decoder, and GRU, demonstrating superior performance for textual attribute imputation.

Unlike general-purpose LLM-based prompting approaches, these transformer models are specifically trained or adapted for structured data, which positions them as a related but distinct research direction.

\subsection{Limitations of Existing Literature}

Across these four paradigms, several recurring limitations emerge:

\begin{itemize}
\item Many prompt-based approaches require iterative querying for each missing value, raising scalability and computational efficiency concerns for large-scale tabular datasets;

\item Existing works frequently focus on a single LLM architecture or a narrow family of models, limiting cross-model comparisons;

\item Evaluations are often conducted on small or domain-specific datasets, with limited benchmarking against strong statistical and machine learning baselines under consistent experimental protocols;

\item Cross-study comparisons remain challenging due to heterogeneous experimental setups and inconsistent treatment of missingness mechanisms (MCAR, MAR, MNAR).
\end{itemize}

In contrast to prior studies, this work provides a systematic and large-scale evaluation of multiple LLM families within a unified tabular imputation framework. Rather than proposing a domain-specific architecture or relying solely on prompt-based enrichment, we conduct a controlled comparative analysis spanning five distinct LLM architectures under standardized missingness mechanisms and experimental conditions.

To the best of our knowledge, no previous study has comprehensively assessed the robustness and comparative behavior of diverse LLMs for tabular missing data imputation within a reproducible and model-agnostic evaluation protocol. All code associated with this study is publicly available via a GitHub repository (Section \textit{Code availability}).

\section{Methodology} \label{sec:methodology}

\begin{figure}[t]
    \centering
    \includegraphics[width=1\textwidth]{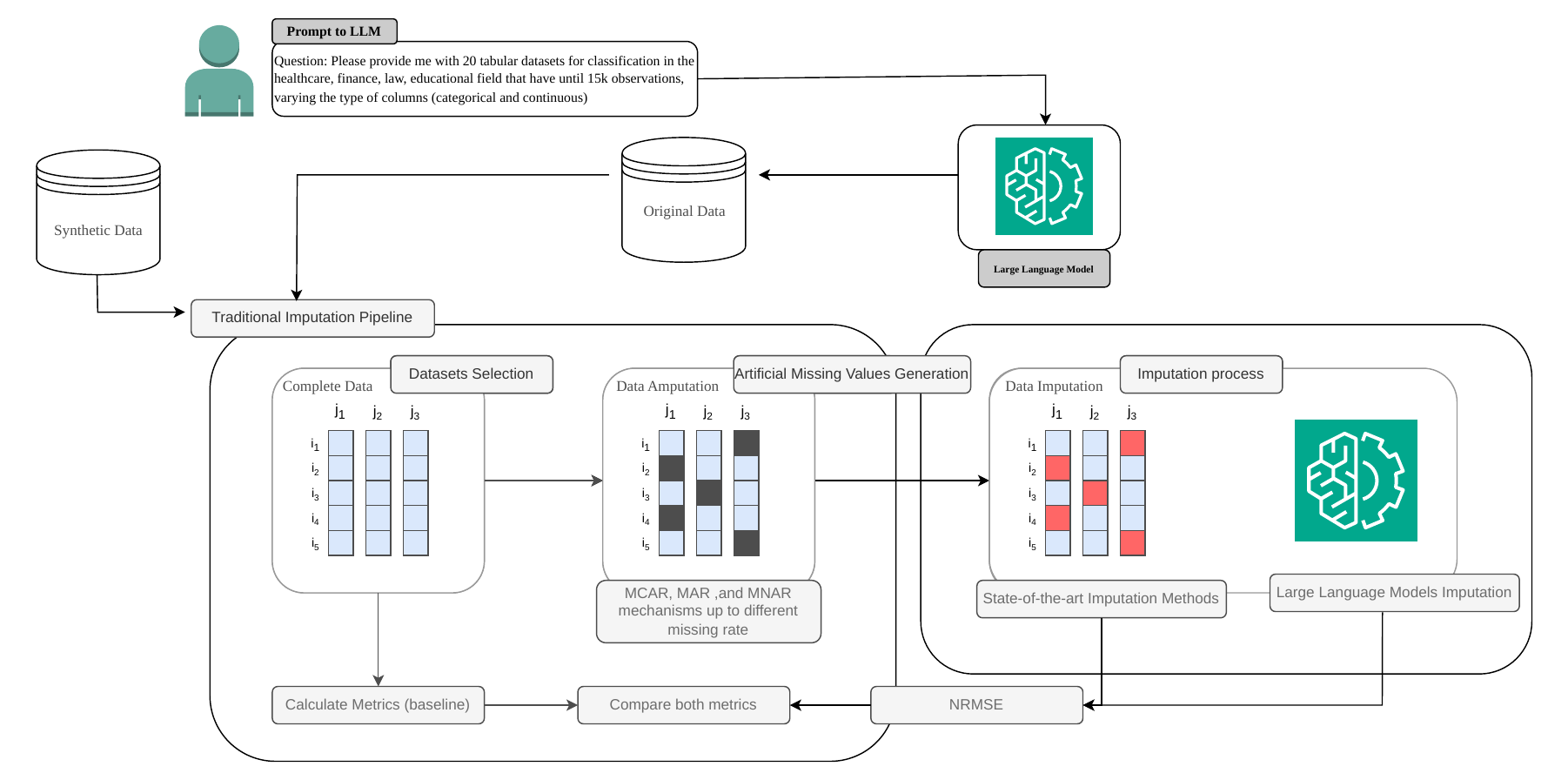}
    \caption{Overview of methodology applied in this work.}
    \label{fig:methodology}
\end{figure}

In this work, we followed the four traditional steps commonly used in benchmarking studies for missing data imputation. Accordingly, Figure~\ref{fig:methodology} provides an overview of the methodology adopted, which is divided into the following stages: Data Collection, Missing Data Synthetic Generation, Data Imputation, and Evaluation Criteria. Each of these steps is detailed in the subsequent sections.

For the experimental setup implementation, we used Python version
3.11.9. All experiments were conducted on a machine equipped with CPU 48 GB of DDR4 RAM (2133 MHz), and running Ubuntu Linux version 22.04.4.

\subsection{Data Collection}

As illustrated in Figure~\ref{fig:methodology}, this step involves two different data sources: synthetic data and datasets derived from LLM background knowledge. For the synthetic data, we used the \texttt{make\_classification} function from the Scikit-learn Python package~\cite{scikit-learn}. This procedure was applied to generate nine different datasets, varying the parameters to this generations, such \texttt{n\_repeated, class\_sep, n\_clusters\_per\_class}, as shown in Table~\ref{tab:synthetic data}. In order to enhance transparency, all these data are freely available in a GitHub repository. Herewith, our goal is to generate data that LLM does not have prior background knowledge during its training. 

\begin{table}[ht]
\caption{Overview of synthetic dataset characteristics.}
\vspace{0.5cm}
\centering
\label{tab:synthetic data}
\resizebox{\textwidth}{!}{
\begin{tabular}{lccc}
\toprule
\multicolumn{1}{c}{\textbf{Dataset Name}}  & \textbf{Samples} & \textbf{Features} & \textbf{Feature Type} \\ \midrule
synthetic-cat & 500 & 4 & Categorical \\
synthetic-cont & 500 & 4 & Continuous \\
synthetic-cat-cont & 500 & 4 & Categorical/Continuous \\ 
synthetic-one & 381 & 4 & Continuous \\
synthetic-two & 131 & 8 & Continuous \\
synthetic-three & 342 & 11 & Continuous \\
synthetic-repeated & 615 & 11 & Continuous \\
synthetic-repeated-two & 649 & 8 & Continuous \\
synthetic-repeated-three & 1000 & 15 & Continuous \\
\bottomrule
\end{tabular}%
}
\end{table}

To identify datasets that are likely represented in the background knowledge of LLMs, we used the Gemini model to assist in the dataset selection process. Gemini was chosen because it is an well-known LLM, which facilitates the reproducibility of the proposed methodology. Specifically, we prompted the model with the following query: \textit{``Please provide me with 20 tabular datasets for classification in the healthcare, finance, law, and education domains with up to 15k observations, varying the types of columns (categorical and continuous)."} The rationale behind this strategy is that datasets suggested by the model are likely to be part of its pretraining corpus or to appear frequently in publicly available internet sources. Consequently, these datasets represent scenarios in which the LLM may already possess semantic or statistical background knowledge. The selected datasets are presented in Table~\ref{tab:uci} and are publicly available in the UCI and/or OpenML repositories.

In this manner, our experimental setup is composed by 29 datasets, 9 synthetic and 20 benchmarks from open-source repositories.

\begin{table}[ht]
\caption{Overview of the datasets characteristics selected by the Gemini model.}
\vspace{0.5cm}
\centering
\label{tab:uci}
\resizebox{\textwidth}{!}{
\begin{tabular}{lcccc}
\toprule
\multicolumn{1}{c}{\multirow{2}{*}{\textbf{Dataset}}} & \multirow{2}{*}{\textbf{Instances}} & \multicolumn{2}{c}{\textbf{Features}} & \multirow{2}{*}{\textbf{\begin{tabular}[c]{@{}c@{}}Subject \\ Area\end{tabular}}} \\ \cmidrule{3-4}
\multicolumn{1}{c}{} &  & \textbf{Continuous} & \textbf{Categorical} &  \\ \midrule
pima-diabetes & 768 & 8 & 1 & Healthcare \\
heart-cleveland & 303 & 6 & 8 & Healthcare \\
wiscosin & 569 & 31 & 1 & Healthcare \\
parkinsons & 195 & 22 & 1 & Healthcare \\
hepatitis & 80 & 6 & 14 & Healthcare \\
mathernal-risk & 1014 & 1 & 7 & Healthcare \\
chronic & 400 & 11 & 14 & Healthcare \\
stalog-heart & 270 & 8 & 7 & Healthcare \\
cervical & 858 & 26 & 6 & Healthcare \\
iris & 150 & 1 & 4 & Biology \\
wine & 178 & 13 & 1 & Physics \\
bc-coimbra & 116 & 9 & 1 & Healthcare \\
student-math & 395 & 16 & 17 & Education \\
student-port & 649 & 16 & 17 & Education \\
user & 403 & 5 & 1 & Education \\
credit-approval & 690 & 6 & 10 & Finance \\
german-credit & 1000 & 7 & 14 & Finance \\
compass-recid & 6172 & 14 & 37 & Law \\
compass-viol-recid & 4020 & 14 & 37 & Law \\
stroke & 5110 & 7 & 4 & Healthcare \\ \bottomrule
\end{tabular}%
}
\end{table}

To introduce artificial missing values (i.e., to perform the missing data synthetic generation process), we adopted a stratified five-fold cross-validation scheme. At each fold, the training data were first normalized to the range [0,1]. Subsequently, for each iteration, we used the Python package \texttt{mdatagen} \cite{mdatagen2025} to generate artificial missing data under three distinct mechanism (MCAR, MAR, and MNAR) in a multivariate setting (i.e., with more than one feature containing missing values). These missing values were introduced independently into the training and test sets to maintain consistency across both sets.

In the MCAR mechanism, every data point has an equal probability of being removed, resulting in a completely random missingness pattern. In contrast, the MAR mechanism was implemented by creating pairs of correlated features, where the feature most correlated with the target variable ($\mathbf{X_{obs}}$) determines the missingness pattern in another feature ($\mathbf{X_{miss}}$), with lower $\mathbf{X_{obs}}$ values yielding lower missingness probabilities. Meanwhile, the MNAR mechanism was implemented by setting the highest missingness probabilities for higher values for each feature~\cite{Santos2019}.

For all three MD scenarios, three different levels of missingness were considered: 5\%, 10\%, 20\% \cite{Ricardo2024SAEI}.
 
\subsection{Imputation Process}

For the imputation process, we selected five state-of-the-art methods: 
$k$NN, MICE~\cite{MICE}, missForest~\cite{missForest}, SAEI~\cite{Ricardo2024SAEI}, and SoftImpute~\cite{Hastie2014}. In addition, we explore TabPFN~\cite{hollmann2025tabpfn} for the imputation task, which represents a novel application of this model within the missing data literature. These approaches are treated as traditional baselines, encompassing both statistical and machine learning-based imputation methods. Their performance serves as a reference for comparison against the following LLMs: Xiaomi MiMo-V2-Flash, Mistral Devstral 2 2512, Gemini 3.0 Flash, GPT-4.1 Nano, and Claude Sonnet-4.5. A comparative analysis of LLMs and cost structures is shown in Table~\ref{tab:llms-overview}.

\begin{table}[htb]
\caption{Comparative analysis of LLMs and cost structures. Pricing is denoted in USD per million tokens (Mtk). Asterisks (*) denote tier-dependent pricing or free-tier availability.}
\vspace{0.5cm}
\label{tab:llms-overview}
\resizebox{\textwidth}{!}{
\begin{tabular}{llcccc}
\toprule
\multicolumn{1}{c}{\multirow{2}{*}{\textbf{Model Identifier}}} & \multicolumn{1}{c}{\multirow{2}{*}{\textbf{API String}}} & \textbf{Input} & \textbf{Output} & \multirow{2}{*}{\textbf{Provider}} & \multirow{2}{*}{\textbf{Access}} \\
\multicolumn{1}{c}{} & \multicolumn{1}{c}{} & \scriptsize\$/Mtk & \scriptsize\$/Mtk &  &  \\ \midrule
MiMo-V2-Flash & \texttt{xiaomi/mimo-v2-flash} & 0.09 & 0.29 & Xiaomi & OpenRouter \\
Devstral 2 2512 & \texttt{mistralai/devstral-2512} & 0.40 & 2.00 & Mistral & OpenRouter \\
Gemini 3.0 Flash & \texttt{gemini-3-flash} & 0.00* & 0.00* & Google & AI Studio \\
GPT-4.1-Nano & \texttt{openai/gpt-4.1-nano} & 0.10 & 0.40 & OpenAI & OpenRouter \\
Claude 4.5 Sonnet & \texttt{claude-sonnet-4.5} & 3.00 & 6.00 & Anthropic & Anthropic \\ \bottomrule
\end{tabular}
}
\end{table}

The implementations of $k$NN, missForest, and MICE were directly obtained from the scikit-learn library~\cite{scikit-learn}, whereas SAEI\footnote{\url{https://github.com/ricardodcpereira/SAEI}}, SoftImpute\footnote{\url{https://github.com/travisbrady/py-soft-impute}}, and TabPFN\footnote{\url{https://github.com/PriorLabs/TabPFN}} were implemented using their respective official GitHub repositories. The hyperparameter configurations adopted for each method are reported in Table~\ref{tab:parameters_imputers}. These settings were selected based on prior literature, including \cite{Mangussi2024Adv,Ricardo2024SAEI}. For TabPFN, as this work is the first to explore this model in data imputation, we have used the default parameters as recommended by the authors~\cite{hollmann2025tabpfn}.

\begin{table}[ht]
\centering
\caption{Parameters of each imputation method.}
\vspace{0.5cm}
\resizebox{0.9\textwidth}{!}{
\label{tab:parameters_imputers}
\begin{tabular}{ll}
\toprule
\multicolumn{1}{c}{\textbf{Imputation method}} & \multicolumn{1}{c}{\textbf{Parameters}} \\ \midrule
$K$NN  & k = 5 (neighbors), Euclidean distance metric \\ 
MICE & 100 iterations, default parameters from Scikit-learn \\ 
SAEI & epochs = 200,  batch size = 64, optimizer = ``Adam"\\ 
missForest & criterion = ``absolute\_error", n\_estimators=10, \\  & default parameters from Scikit-learn\\ 
TabPFN & default parameters\\

\bottomrule
\end{tabular}}
\end{table}

For the LLM-based methods, the temperature parameter was fixed at 0.1 to encourage deterministic outputs. Moreover, for Gemini, GPT, and Claude models, external search or tool-augmented capabilities were disabled, ensuring that only the prompt content was used as input during the imputation process.

Since this study focuses on evaluating the effectiveness of prompt engineering to address our research hypotheses, tabular data were provided to the LLMs in batch form within the prompt. Section~\ref{subsec:prompt} details the batch construction strategy and its role in the imputation process. Due to the stochastic nature of prompt-based inference in LLMs, variations in output structure may occur across different requests. To ensure robustness and reproducibility, our pipeline incorporates explicit fallback mechanisms to handle potential inconsistencies and API-related failures.

Specifically, if an LLM fails to return a valid imputed batch, either due to mismatches in the expected dataframe shape (e.g., incorrect number of columns), the presence of invalid values such as \texttt{NaN}, formatting inconsistencies, or incompatible feature representations, the request is automatically retried up to three times. Each retry follows an exponential backoff strategy, progressively increasing the delay between attempts to mitigate temporary API unavailability, rate limiting, or overload issues. If, after three attempts, the LLM still fails to generate a structurally valid imputation output, a deterministic fallback strategy is applied. In line with common practices in the missing data literature for the missForest method, as described in the survey by Sun et al.~\cite{Sun2023Deep}, a preliminary statistical imputation step (specifically mean imputation) is applied to ensure that the dataset remains complete and suitable for downstream evaluation. This robustness protocol guarantees experimental continuity while preventing stochastic API behavior from biasing comparative performance results.

For all models evaluated in our experimental setup, we employed stratified five-fold cross-validation. For non-LLM-based methods, the input data were normalized to the [0,1] range. In contrast, data normalization was not applied for LLM-based approaches. This design choice aligns with our hypothesis that LLMs, owing to their extensive pre-training on large-scale and diverse corpora, may inherently handle raw numerical values without explicit normalization. 

To evaluate the quality of the imputation process, we employ the Normalized Root Mean Square Error (NRMSE), as defined in Equation~\ref{eq:nrmse}. In this formulation, $N$ denotes the number of observations, $\hat{y}_i$ represents the imputed value for the 
i-th observation, and $y_i$ corresponds to the ground-truth value. The terms $y_{\max}$ and $y_{\min}$ denote the maximum and minimum values of the corresponding feature, respectively. Lower NRMSE values indicate better imputation performance, with values closer to zero reflecting more accurate imputations.

\begin{equation}
\label{eq:nrmse}
\mathrm{NRMSE} = \frac{\sqrt{\frac{1}{N} \sum_{i=1}^{N} (\hat{y}_i - y_i)^2}}{y_{\max} - y_{\min}}
\end{equation}

In addition, to complement the evaluation of imputation quality, we consider computational efficiency metrics in our experimental setup. Specifically, we measure the computational time, the number of tokens processed, and the associated input and output costs to assess the practical feasibility of employing LLMs for data imputation.

\subsection{Prompt Construction and Batch Strategy} \label{subsec:prompt}

\begin{figure}[h]
\centering
\includegraphics[width=\linewidth]{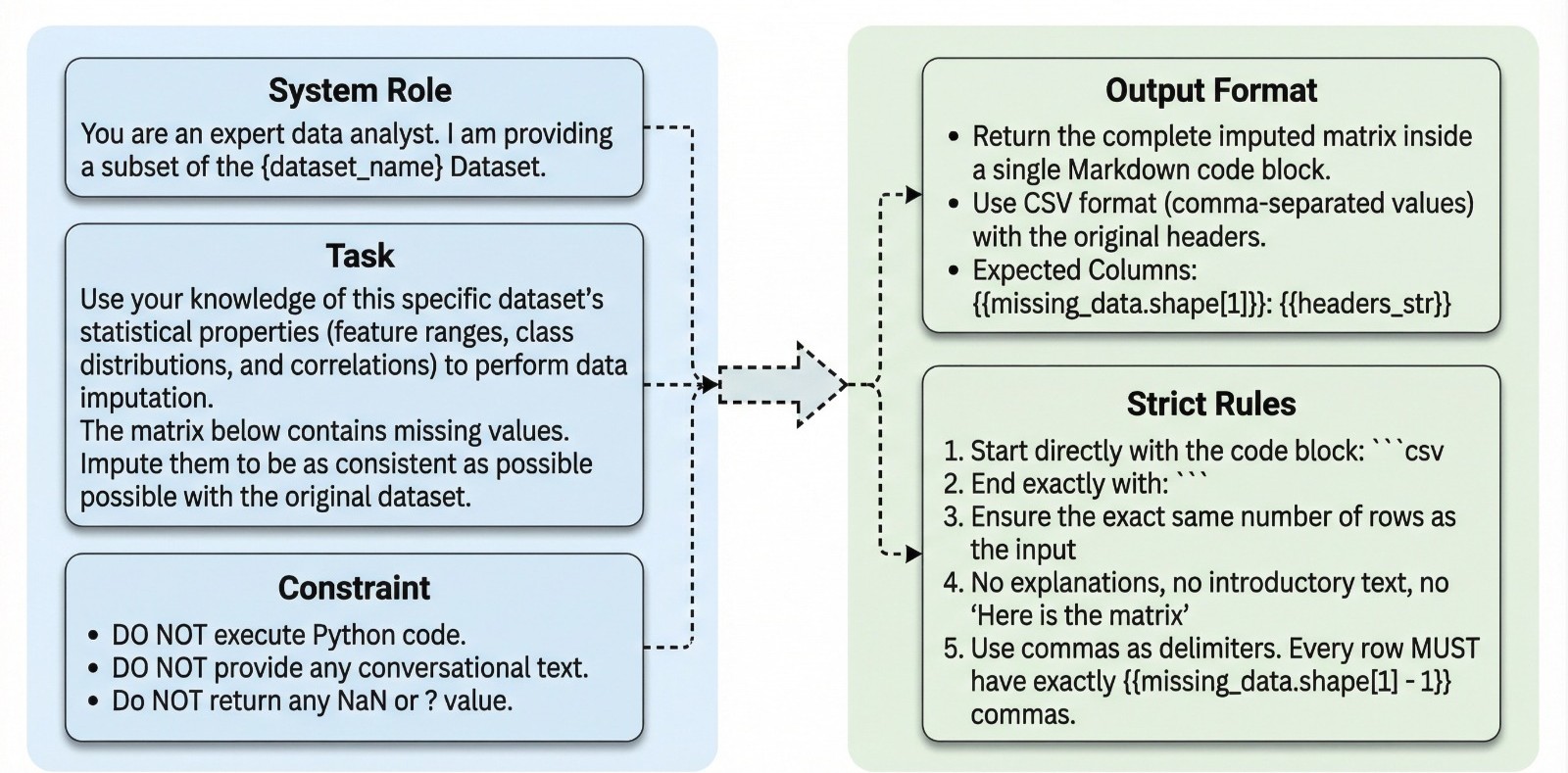}
\caption{Illustration of the complete prompt structure used to perform data imputation via prompt engineering.}
\label{fig:prompt}
\end{figure}

An empirical study was conducted to derive the final prompt used for the data imputation task, as illustrated in Figure~\ref{fig:prompt}. First, the system role (persona) was defined to frame the LLM as a data analyst, thereby contextualizing the task and guiding the model’s responses. Since one of the primary objectives of this work is to explore the background knowledge encoded in LLMs, the introductory sentence \textit{``I am providing a subset of the dataset”} plays a crucial role in informing the model about the nature of the input and encouraging it to leverage prior knowledge when processing the subsequent prompt components.

Initially, the prompt consisted solely of a task description. However, preliminary experiments revealed that this configuration led to unstable and inconsistent outputs across different LLMs. Further investigation, particularly through controlled experiments using the Gemini model in Google AI Studio, revealed that some LLMs attempted to execute implicit computational procedures (e.g., describing or emulating a $k$NN imputer or by mean) or generated verbose reasoning explanations instead of directly imputing missing values.

To address these issues, a Constraints block (see Figure~\ref{fig:prompt}) was incorporated into the prompt. This block explicitly restricted the models from executing code, providing extended explanations, or returning invalid placeholders such as ``\textit{NaN”} or \textit{``?”}. These constraints proved essential for enforcing consistent behavior across models.

Given that our experimental design considers three missing data mechanisms (MCAR, MAR, and MNAR) with missing rates of 5\%, 10\%, and 20\% across multiple LLMs, a fully automated pipeline was developed to execute the proposed methodology. This pipeline is publicly available in the associated GitHub repository. During initial executions, several runtime and formatting errors were observed, particularly due to variations in model outputs. To mitigate these issues and ensure robustness, additional Output Format and Strict Rules blocks were incorporated into the prompt, enabling reliable and reproducible imputation results across large-scale experiments.

Figure~\ref{fig:prompt} presents the complete structure of the final prompt used to obtain the experimental results reported in this work. Furthermore, as described in Section~\ref{sec:methodology}, a batch-based strategy was adopted to provide tabular data to the LLMs efficiently within the prompt.

After defining the final prompt, the size of the dataset subset provided to the LLMs, drawn from the datasets listed in Table~\ref{tab:uci}, was fixed at $40 \times 10$ (i.e., 40 observations with 10 features). This configuration was selected based on an empirical evaluation of multiple input sizes, including $20\times5$, $20\times8$, $20\times10$, $40\times10$, $50\times10$, $80\times10$, and $100\times10$. Among these configurations, the $40\times10$ subset achieved the best trade-off between computational efficiency, adherence to prompt constraints, and imputation quality, and was therefore adopted throughout the experimental analysis.

This $40 \times 10$ window was applied in a sliding-window manner across each dataset and each fold of the stratified cross-validation procedure to perform the imputation process.

\section{Results and Discussion} \label{sec:results}

This section presents the main findings for data imputation through LLMs using the prompt engineering approach. Additionally, a comparative analysis between LLMs and traditional methods are outlined. Afterward, a case study in worse-case scenarios was analyzed.

\subsection{Overall Performance}

\begin{figure}[!htb]
    \centering
    \includegraphics[width=\textwidth]{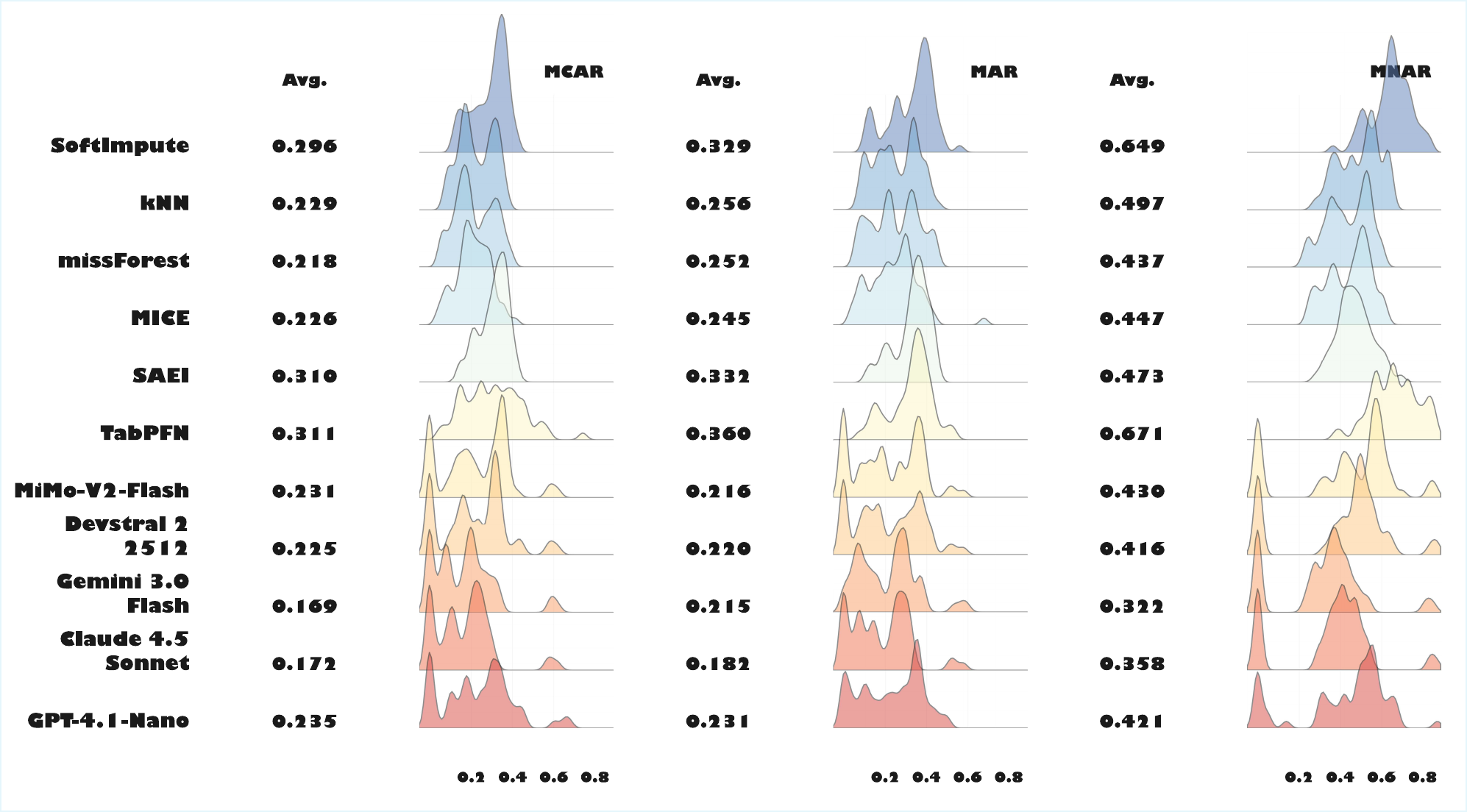}
    \caption{Density plots of the Normalized Root Mean Square Error results obtained for each imputation strategy. The analysis is presented by missing data mechanisms. The average NRMSE for each setting is also displayed.}
    \label{fig:overall}
\end{figure}

As previously mentioned, the experimental setup compares five different LLMs for the data imputation task. An overview of how these models are accessed, including their providers and API endpoints, is presented in Table~\ref{tab:llms-overview}. Based on this configuration, Figure~\ref{fig:overall}  presents the density plots of the NRMSE results obtained for each imputation strategy. These plots considering all real-world datasets and missing rates under MCAR, MAR, and MNAR mechanisms.

This initial finding addresses \textbf{RQ1}, showing that different models using the same prompt can exhibit substantial performance differences. Although the overall objective is achieved across all five evaluated models, the Xiaomi, Mistral, and GPT do not reach performance levels comparable to those of Gemini 3.0 Flash and Claude 4.5 Sonnet. Conversely, GPT-4.1-Nano represents the worst-case scenario for the data imputation task given the LLM architecture. These results indicate that proprietary access does not necessarily translate into superior imputation performance. 

We hypothesize that the training processes of Gemini 3.0 Flash and Claude 4.5 Sonnet, as the most recent models evaluated in this study, contribute substantially to the strong performance observed in Figure~\ref{fig:overall}. According to publicly available documentation, the training cut-off dates for Gemini 3.0 Flash and Claude 4.5 Sonnet are January 2025 and July 2025, respectively. These models are designed as general-purpose systems for a wide range of tasks, including conversational assistance, code generation, and multimodal reasoning, and are therefore trained on large-scale data collected from diverse public sources, including websites and open repositories, with additional data used during post-training and continual refinement.

Given the increasing use of open-source repositories and publicly available datasets in AI research and model training, it is plausible that the datasets listed in Table~\ref{tab:uci}, which are widely used and publicly accessible, may already be represented—directly or indirectly—in the training corpora of these models. This prior exposure could partially explain their superior performance relative to other LLMs, leading to overall optimistic results when compared to traditional imputation methods on real-world datasets.

To assess the statistical significance of the observed differences, a Three-Way ANOVA on ranks was conducted at a 5\% significance level. The factors considered were the dataset, the missing rate, and the imputation method, with NRMSE as the dependent variable. Since the assumptions of normality required for a standard Three-Way ANOVA were violated based on Anderson Darling test, the data were transformed using Ordered Quantile normalization, which yields normally distributed transformed values. The same assumptions were verified for all data subgroups. The resulting p-values indicate that all factors are statistically significant: both the dataset and imputation method factors yield $p<2e^{-16}$, respectively , while the missing rate factor yields $p=0.0221$.

Finally, to assess whether LLM-based imputation methods significantly outperformed the remaining approaches at a 5\% significance level, a post-hoc Tukey’s HSD test was applied considering the imputation method factor. It revealed that Gemini 3.0 Flash and Claude 4.5 Sonnet significantly outperformed all classical imputation methods $(p < 0.01)$. No statistically significant difference was observed between Gemini 3.0 Flash and Claude 4.5 Sonnet $(p = 1.0)$, indicating equivalent performance. Classical methods such as kNN, missForest, and MICE did not differ significantly among themselves $(p=0.9171, 0.9491822, 1.000)$, respectively.

\subsection{Degradation Under Increasing Missing Rates}

To further analyze the behavior of LLM-based imputation, the NRMSE was evaluated as a function of the missing rate under the MNAR, MCAR, and MAR mechanisms. Table~\ref{tab:missing_rates} reports the average NRMSE across different missing rates in our experimental design, where bold values indicate the best-performing method and underlined values denote the second best.

Consistent with the overall results, Gemini 3.0 Flash and Claude 4.5 Sonnet emerge as the two top-performing models under the MNAR and MCAR mechanisms. However, under the MAR mechanism, Claude 4.5 Sonnet achieved the best performance across all missing rates. Regarding the second-best results, Mistral ranked second at 5\% and 10\% missing rates, while Gemini 3.0 Flash ranked second at 20\%.

As expected, increasing the missing rate leads to higher imputation error, as measured by NRMSE, a trend well established in the missing data literature for traditional imputation methods~\cite{Mangussi2024Adv}. Contrary to our initial hypothesis, however, the degradation in performance with increasing missingness is less pronounced for LLM-based approaches. This suggests that, despite higher missing rates, LLMs are able to leverage contextual and semantic relationships among features to support the imputation process. 

% \usepackage{graphicx}
% \usepackage[normalem]{ulem}
\begin{table}[]
\caption{Overall Normalized Root Mean Square Error (NRMSE) divided by missing rate on real-world datasets under MNAR, MCAR, and MAR mechanisms for all large language models. Bold values indicate the best-performing method, while underlined values denote the second best.}
\vspace{0.5cm}
\label{tab:missing_rates}
\resizebox{\textwidth}{!}{%
\begin{tabular}{llllllllll}
\hline
\multicolumn{1}{c}{\textbf{MD   Mechanisms}} & \multicolumn{3}{c}{\textbf{MNAR}} & \multicolumn{3}{c}{\textbf{MCAR}} & \multicolumn{3}{c}{\textbf{MAR}} \\ \hline
\multicolumn{1}{c}{\textbf{Missing Rate}} & \multicolumn{1}{c}{\textbf{5\%}} & \multicolumn{1}{c}{\textbf{10\%}} & \multicolumn{1}{c}{\textbf{20\%}} & \multicolumn{1}{c}{\textbf{5\%}} & \multicolumn{1}{c}{\textbf{10\%}} & \multicolumn{1}{c}{\textbf{20\%}} & \multicolumn{1}{c}{\textbf{5\%}} & \multicolumn{1}{c}{\textbf{10\%}} & \multicolumn{1}{c}{\textbf{20\%}} \\ \hline
SoftImpute & 0.654 & 0.644 & 0.649 & 0.273 & 0.294 & 0.320 & 0.311 & 0.325 & 0.351 \\
kNN & 0.485 & 0.496 & 0.509 & 0.203 & 0.228 & 0.256 & 0.236 & 0.249 & 0.284 \\
missForest & 0.418 & 0.440 & 0.453 & 0.192 & 0.218 & 0.242 & 0.233 & 0.242 & 0.283 \\
MICE & 0.426 & 0.439 & 0.475 & 0.174 & 0.212 & 0.292 & 0.211 & 0.227 & 0.298 \\
SAEI & 0.518 & 0.482 & 0.418 & 0.295 & 0.313 & 0.320 & 0.330 & 0.333 & 0.335 \\
TabPFN & 0.621 & 0.683 & 0.710 & 0.219 & 0.276 & 0.437 & 0.317 & 0.354 & 0.411 \\
Xiaomi: MiMo-V2-Flash & 0.439 & 0.435 & 0.416 & 0.207 & 0.236 & 0.249 & 0.204 & 0.221 & 0.225 \\
Mistral: Devstral 2 2512 & 0.435 & 0.424 & 0.389 & 0.210 & 0.229 & 0.236 & {\ul 0.207} & {\ul 0.218} & 0.235 \\
Gemini 3.0 Flash & \textbf{0.333} & \textbf{0.325} & \textbf{0.308} & \textbf{0.150} & \textbf{0.172} & \textbf{0.185} & 0.211 & 0.234 & {\ul 0.200} \\
Claude 4.5 Sonnet & {\ul 0.369} & {\ul 0.361} & {\ul 0.345} & {\ul 0.153} & {\ul 0.175} & {\ul 0.188} & \textbf{0.168} & \textbf{0.182} & \textbf{0.196} \\
Gpt - 4.1-Nano & 0.432 & 0.405 & 0.425 & 0.221 & 0.234 & 0.252 & 0.221 & 0.232 & 0.240 \\ \hline
\end{tabular}%
}
\end{table}

Finally, the results presented in Figure~\ref{fig:overall} indicate that the MNAR scenario is the most challenging missing data mechanism, which is consistent with prior findings in the literature. Additionally, LLM-based imputation exhibits similar performance under the MAR and MCAR mechanisms. In a real-world context, these results point toward potentially impactful analytical opportunities. For example, in healthcare, LLMs could leverage semantic context from medical reports to infer missing values. Given the large-scale knowledge encoded across diverse scenarios, the explanatory capacity of LLMs may be particularly valuable for imputation under MNAR settings, where missingness depends on unobserved values. In such cases, LLMs can leverage their ability to extrapolate from contextual and semantic patterns learned from internet-scale corpora.

\subsection{LLM-Based vs. Traditional Imputation}

With respect to \textbf{RQ2}, Gemini 3.0 Flash and Claude 4.5 Sonnet outperform traditional methods in the data imputation task, as shown in Figure~\ref{fig:overall}. These results demonstrate that these LLMs surpass state-of-the-art imputation methods, such as MICE under the MAR mechanism. Based on the findings shown, our analysis focuses on scenarios in which at least one Large Language Model outperforms traditional imputation methods across different missing rates under the MCAR, MAR, and MNAR mechanisms. These scenarios are summarized in Table~\ref{tab:scenarios}.

\begin{table}[ht]
\small
\caption{Scenarios where at least one Large Language Model achieves superior performance compared to traditional data imputation methods.}
\vspace{0.5cm}
\label{tab:scenarios}
\resizebox{\textwidth}{!}{ % This is the secret to fixing the hbox error
\begin{tabular}{l ccc ccc ccc }
\hline
\multicolumn{1}{c}{\multirow{2}{*}{\textbf{Dataset}}} & \multicolumn{3}{c}{\textbf{5\%}} & \multicolumn{3}{c}{\textbf{10\%}} & \multicolumn{3}{c}{\textbf{20\%}} \\ \cmidrule{2-10} 
\multicolumn{1}{c}{} & \textbf{MNAR} & \textbf{MCAR} & \textbf{MAR} & \textbf{MNAR} & \textbf{MCAR} & \textbf{MAR} & \textbf{MNAR} & \textbf{MCAR} & \textbf{MAR} \\ \hline
pima & \checkmark & \checkmark & - & \checkmark & \checkmark & \checkmark & \checkmark & - & \checkmark \\
cleveland & \checkmark & \checkmark & \checkmark & \checkmark & \checkmark & \checkmark & \checkmark & \checkmark & \checkmark \\
wiscosin & - & - & - & - & - & - & - & - & - \\
parkinsons & \checkmark & \checkmark & \checkmark & \checkmark & \checkmark & \checkmark & \checkmark & \checkmark & \checkmark \\
hepatitis & - & - & - & - & - & - & - & - & - \\
mathernal-risk & - & - & - & \checkmark & - & - & \checkmark & - & \checkmark \\
chronic & - & \checkmark & \checkmark & - & \checkmark & \checkmark & \checkmark & \checkmark & \checkmark \\
stalog & \checkmark & - & - & \checkmark & - & - & \checkmark & - & \checkmark \\
cervical & - & \checkmark & \checkmark & \checkmark & \checkmark & \checkmark & \checkmark & \checkmark & \checkmark \\
iris & \checkmark & \checkmark & \checkmark & \checkmark & \checkmark & \checkmark & \checkmark & \checkmark & \checkmark \\
wine & \checkmark & \checkmark & \checkmark & \checkmark & \checkmark & \checkmark & \checkmark & \checkmark & \checkmark \\
bc-coimbra & \checkmark & \checkmark & \checkmark & \checkmark & \checkmark & \checkmark & \checkmark & \checkmark & \checkmark \\
student-math & - & - & - & \checkmark & - & \checkmark & \checkmark & \checkmark & \checkmark \\
student-port & - & \checkmark & - & \checkmark & \checkmark & \checkmark & \checkmark & \checkmark & \checkmark \\
user & \checkmark & \checkmark & \checkmark & \checkmark & \checkmark & \checkmark & \checkmark & \checkmark & \checkmark \\
credit-approval & \checkmark & \checkmark & \checkmark & \checkmark & \checkmark & \checkmark & \checkmark & \checkmark & \checkmark \\
german-credit & \checkmark & - & - & \checkmark & - & - & \checkmark & - & - \\
compass-4k & \checkmark & - & - & - & - & - & \checkmark & - & - \\
stroke & \checkmark & \checkmark & - & - & \checkmark & - & \checkmark & \checkmark & \checkmark \\
compass-7k & - & - & - & \checkmark & - & - & - & - & \checkmark \\ \hline
\end{tabular}%
}
\end{table}

As shown in the results, the \textit{Heart Disease (Cleveland), Parkinson’s Disease (Voice), Iris, Wine, Breast Cancer Coimbra, User Knowledge Modeling,} and \textit{Credit Card Approvals} datasets represent the most favorable scenarios for LLM-based imputation. In contrast, the remaining datasets indicate that LLMs face greater challenges in outperforming traditional imputation methods. Notably, the \textit{Hepatitis, Breast Cancer Wisconsin (Diagnostic), German Credit, COMPAS Recidivism,} and \textit{COMPAS Violent Recidivism} datasets constitute worst-case scenarios, where traditional approaches consistently outperform all evaluated LLMs.

These findings reveal a meaningful pattern. The variance in LLM performance across datasets suggests a strong dependency on semantic density and pre-training alignment. LLMs demonstrate superior imputation capabilities in datasets such as \textit{Heart Disease (Cleveland)} and \textit{Iris}, where they are very applied across internet-corpora.

Conversely, the poorer performance observed in datasets such as \textit{German Credit} and \textit{COMPAS} may be explained by one primary factor: high dimensionality. Unlike traditional statistical methods (e.g., MICE or missForest), which rely purely on numerical variance and conditional modeling, LLMs depend heavily on semantic signals.As dimensionality increases, the self-attention mechanism may become diluted across long input sequences, making it more difficult to capture complex, high-dimensional feature interactions that tree-based or iterative statistical methods can exploit effectively. Thus, while LLMs offer a promising reasoning-oriented paradigm for missing data imputation, their performance remains constrained by feature interpretability and structural complexity.

The Hepatitis dataset is relatively small ($80 \times 20$). Under the adopted five-fold stratified cross-validation design, each fold contains approximately 15 observations. With a batch size of $40 \times 10$, this configuration results in two batches per fold. As discussed in Section~\ref{subsec:prompt}, smaller prompt kernels negatively impact LLM imputation performance. Furthermore, the average prompt length for \textit{Hepatitis} is 1,047 tokens, roughly half the prompt size used for the \textit{Parkinson’s Disease (Voice)} dataset. This suggests that shorter prompts may provide insufficient contextual information for accurate imputation.

In contrast, the \textit{Breast Cancer Wisconsin (Diagnostic)} dataset has an average prompt length of approximately 2,000 tokens, indicating that prompt size alone does not explain its lower performance. Instead, this dataset exhibits the highest dimensionality among all evaluated datasets and requires nine batches to impute a single fold. This fragmentation across multiple batches likely disrupts global contextual coherence, limiting the LLM’s ability to model inter-feature dependencies effectively.

In summary, LLM-based imputation in batches appears highly sensitive to both prompt structure and dataset characteristics. Small sample sizes combined with short prompts (as in \textit{Hepatitis}) restrict contextual information, while high dimensionality and excessive batch fragmentation (as in \textit{Breast Cancer Wisconsin (Diagnostic)}) impair the preservation of global dependencies. These results indicate that LLM-based imputation performs best when contextual coherence can be maintained within prompts, emphasizing the importance of carefully designed batching strategies and prompt configurations for complex tabular data.

\subsection{Hallucinations and Fallback Mechanism}

Regarding \textbf{RQ3}, not all LLMs appear to be equally prone to hallucinations in imputation tasks. Although the overall results for real-world datasets are encouraging, further analysis from a hallucination perspective provides additional insights.

To investigate this aspect, we analyzed the activation of the fallback mechanism, which is triggered when the model fails to produce a valid imputation. The percentage of cases in which LLMs relied on the fallback mechanism is reported in Table~\ref{tab:fallback}. These results were measured using synthetic datasets, as LLM-based methods do not possess prior knowledge about these artificially generated data distributions, making this setting more susceptible to hallucinations.

As shown in Table~\ref{tab:batch}, GPT-4.1-Nano did not consistently follow the zero-shot prompt instructions, which resulted in several activations of the fallback mechanism. This behavior suggests that more compact models may be more susceptible to hallucinations or prompt-following failures in imputation tasks. In contrast, Claude 4.5 Sonnet and Gemini 3.0 Flash never required the fallback mechanism. Moreover, these models showed no evidence of hallucinations in the imputed values, demonstrating consistently strong performance across the evaluated experimental scenarios. These findings remain consistent across the entire experimental setup.

% \usepackage{graphicx}
\begin{table}[htb]
\caption{Comparative analysis of the percentage of cases in which LLMs relied on the fallback mechanism for missing data imputation. The results are based on synthetic datasets, where LLM-based methods exhibited poorer performance.}
\centering
\vspace{0.5cm}
\label{tab:fallback}
\resizebox{0.6\textwidth}{!}{%
\setlength{\tabcolsep}{20pt} 
\begin{tabular}{lc}
\hline
\multicolumn{1}{c}{\textbf{LLMs}} & \textbf{Fallback Usage (\%)} \\ \hline
MiMo-V2-Flash & 0\% \\
Devstral 2 2512 & 1.85\% \\
Gemini 3.0 Flash & 0\% \\
Claude 4.5 Sonnet & 0\% \\
GPT-4.1-Nano & 14.94\% \\ \hline
\end{tabular}%
}
\end{table}

\subsection{Pareto Analysis of Computational Cost and Imputation Accuracy}

\begin{figure}[!htb]
    \centering
    \includegraphics[width=\textwidth]{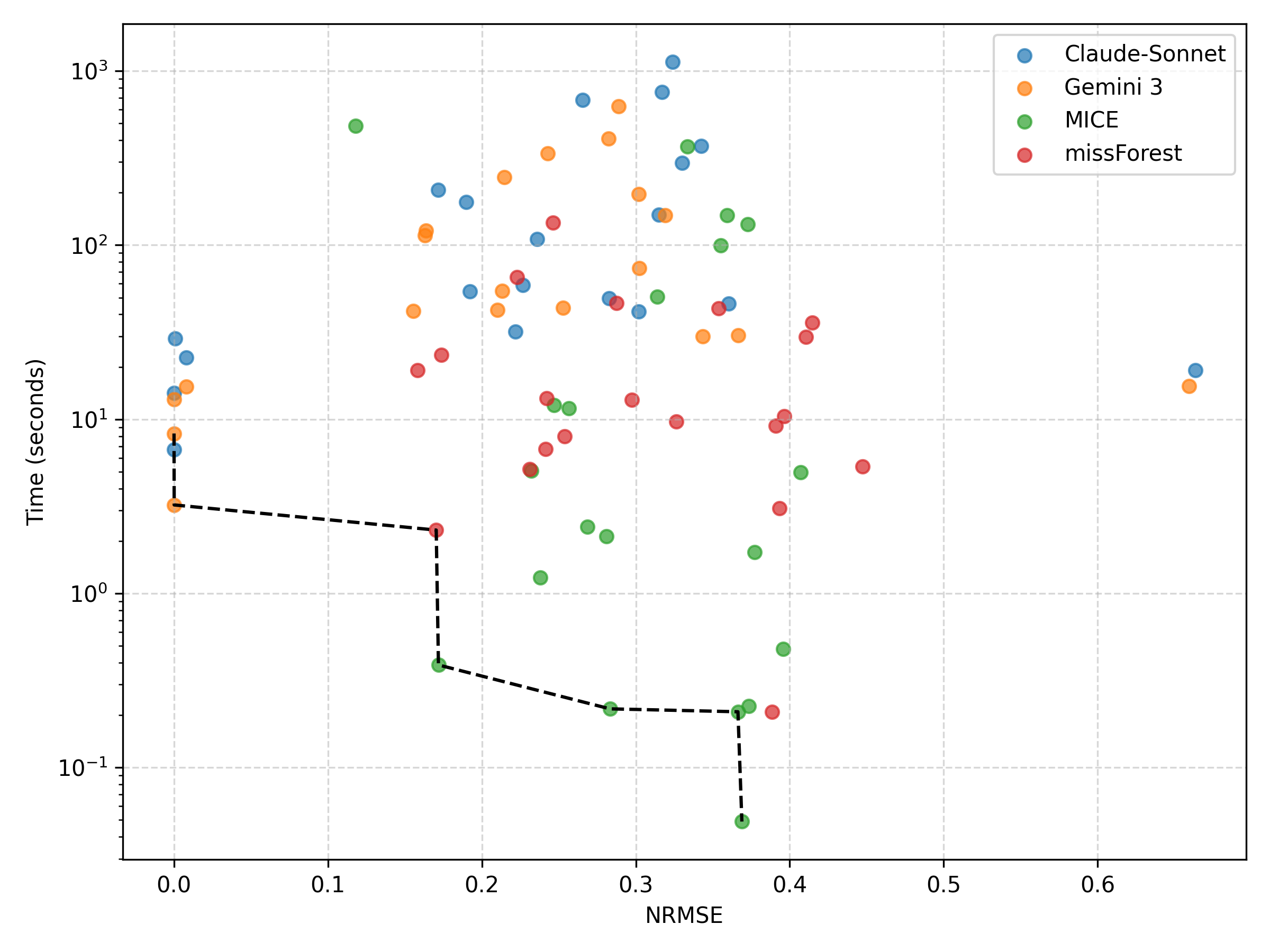}
    \caption{NRMSE vs. computational time across imputation methods.
Each point corresponds to a dataset. The x-axis reports imputation accuracy (NRMSE), while the y-axis shows runtime in seconds (log scale). The figure highlights the trade-off between predictive performance and computational cost across LLM-based and traditional imputation approaches with Pareto frontier.}
    \label{fig:scatter}
\end{figure}

Considering the two best-performing LLM-based imputation methods and the two strongest traditional baselines, Gemini 3.0 Flash, Claude 4.5 Sonnet, MICE, and missForest, we evaluated their computational efficiency. Figure~\ref{fig:scatter} illustrates the trade-off between imputation error (measured by the NRMSE metric) and computational cost, highlighting the Pareto frontier.

As described in Section~\ref{subsec:prompt}, the prompt engineering strategy adopted for LLM-based imputation relies on a batch-wise processing approach, which is not required for traditional imputation methods. Table~\ref{tab:batch} reports the number of batches required for each real-world dataset to complete the imputation task. Since each batch corresponds to an individual API request, the number of batches directly increases the computational cost for Claude 4.5 Sonnet and Gemini 3.0 Flash, leading to a trade-off between imputation accuracy and computational efficiency.

\begin{table}[htb]
\centering
\caption{Number of batches per fold in real-world datasets used for LLMs-based to perform data imputation task.}
\vspace{0.5cm}
\label{tab:batch}
\resizebox{0.5\textwidth}{!}{
% 1. Increase padding slightly so the narrow table looks balanced
\setlength{\tabcolsep}{30pt} 
% 2. Use standard tabular (no width argument) so it stays compact

\begin{tabular}{cc}
\toprule
\textbf{Dataset} & \textbf{NBF} \\ \midrule
pima & 4 \\
cleveland & 4 \\
wiscosin & 9 \\
parkinsons & 3 \\
hepatitis & 2 \\
mathernal-risk & 6 \\
chronic & 3 \\
stalog & 4 \\
cervical & 15 \\
iris & 1 \\
wine & 2 \\
bc-coimbra & 1 \\
student-math & 10 \\
student-port & 20 \\
user & 3 \\
credit-approval & 8 \\
german-credit & 25 \\
compass-4k & 65 \\
stroke & 50 \\
compass-7k & 93 \\ \bottomrule
\end{tabular}
}
\end{table}

As shown in Figure~\ref{fig:scatter}, Claude 4.5 Sonnet and Gemini 3.0 Flash achieve excellent performance on datasets such as \textit{Iris, Wine, Breast Cancer Coimbra}, and \textit{User Knowledge Modeling}. However, the batch-processing strategy substantially increases computational time compared to MICE and missForest. Overall, MICE emerges as the fastest imputation method among the evaluated approaches, as indicated by its position on the Pareto frontier.

It is important to note that the number of batches is constrained by the request limits imposed by the service providers hosting the LLMs. Factors such as platform demand, request throughput, and latency between API calls and response generation contribute significantly to the observed computational cost. An additional limitation of LLM-based methods is the associated cost of both input and output tokens, which is not incurred by traditional techniques.

\subsection{Evaluation Performance of LLMs in Synthetic Data}

Regarding \textbf{RQ2}, \textbf{RQ3}, and the second part of our methodology, the objective is to investigate whether the internet-scale corpora used to pre-train LLMs provide additional value for the missing data imputation task. As described in Section~\ref{sec:methodology}, nine synthetic datasets were generated, each varying in feature types, to enable controlled experimentation.

To examine our hypothesis that the background knowledge encoded in LLMs may lead to overly optimistic results, we compared the performance of Claude 4.5 Sonnet and Gemini 3.0 Flash against two strong traditional baselines, MICE and missForest (the best two-performing methods in real-world datasets in our experimental design), on these synthetic datasets. Table~\ref{tab:synthetic_results} reports the overall imputation performance under the MCAR, MAR, and MNAR missingness mechanisms. The methodology adopted in this experiment follows the same procedure described in Section~\ref{sec:methodology}. The only modification concerns the dataset identifier provided in the prompt, which, in this case, was specified as \textit{Synthetic}.

\begin{table}[]
\caption{Overall Normalized Root Mean Square Error (NRMSE) grouped by missing rate on synthetic datasets under MNAR, MCAR, and MAR mechanisms for Gemini 3.0 Flash, Claude 4.5 Sonnet, MICE, and missForest. Bold values indicate the best-performing method, while underlined values denote the second best.}
\centering
\vspace{0.5cm}
\label{tab:synthetic_results}
\resizebox{0.8\textwidth}{!}{
\begin{tabular}{llll}
\toprule
\multicolumn{1}{c}{\multirow{2}{*}{\textbf{\begin{tabular}[c]{@{}c@{}}Imputation   \\ Methods\end{tabular}}}} & \multicolumn{3}{c}{\textbf{MD Mechanism}} \\ \cmidrule{2-4} 
\multicolumn{1}{c}{} & \multicolumn{1}{c}{\textbf{MAR}} & \multicolumn{1}{c}{\textbf{MNAR}} & \multicolumn{1}{c}{\textbf{MCAR}} \\ \midrule
missForest & {\ul 0.224 $\pm$ 0.122} & {\ul 0.425 $\pm$ 0.096} & {\ul 0.188 $\pm$ 0.140} \\
MICE & \textbf{0.160 $\pm$ 0.146} & 0.429 $\pm$ 0.115 & \textbf{0.169 $\pm$ 0.137} \\
Gemini 3.0 Flash & 0.239 $\pm$ 0.144 & \textbf{0.422 $\pm$ 0.153} & 0.208 $\pm$ 0.134 \\
Claude 4.5 Sonnet & 0.257 $\pm$ 0.134 & 0.433 $\pm$ 0.152 & 0.225 $\pm$ 0.143 \\ \bottomrule
\end{tabular}}
\end{table}

As shown in Table~\ref{tab:synthetic_results}, traditional imputation methods, particularly MICE, consistently outperform all evaluated LLM-based approaches under MCAR and MAR mechanisms on synthetic datasets. Although Gemini 3.0 Flash demonstrates competitive performance relative to other LLM-based imputation methods, it does not surpass classical statistical approaches in these controlled settings. This behavior contrasts sharply with the results observed on real-world datasets, where LLMs—especially Gemini 3.0 Flash and Claude 4.5 Sonnet—achieve superior imputation performance across all missing data mechanisms.

This divergence directly addresses \textbf{RQ2} and \textbf{RQ3}. The strong performance of LLMs on open-source, real-world datasets, combined with their weaker results on synthetic data, suggests that the effectiveness of LLM-based imputation is closely tied to the availability of semantic and contextual information acquired during large-scale pre-training. In contrast, synthetic datasets (by construction) lack real-world semantics and are unlikely to resemble data encountered during pre-training. As a result, the advantage derived from prior knowledge is largely neutralized. Thus, a hallucinations from LLMs is more likely to occur in imputation contexts that are unfamiliar to the model.

The MNAR scenario presents a more nuanced picture. In this setting, Gemini 3.0 Flash remains competitive with MICE and missForest, which is a particularly interesting finding. Since the MNAR mechanism, often introduces complex, non-random patterns, the ability of an LLM to remain competitive under such conditions suggests that reasoning-based imputation may capture certain structured dependencies beyond purely variance-driven modeling. This observation opens an important avenue for further investigation.

Overall, these findings indicate that LLM-based imputation substantially benefits from prior exposure to domain-specific patterns embedded in real-world data, while offering limited advantages in fully controlled synthetic environments. Rather than reflecting a methodological weakness, this behavior reveals a fundamental distinction between semantic-driven imputation and purely statistical reconstruction, highlighting the potentially complementary nature of LLM-based and traditional imputation approaches.

\section{Conclusions} \label{sec:conclusion}

In this work, we investigated the robustness of LLMs for performing data imputation in tabular datasets. Our central hypothesis was that, due to their training on vast internet-scale corpora, LLMs implicitly encode knowledge of many widely used open-source datasets. As a result, these models may leverage such prior knowledge to outperform state-of-the-art imputation methods under comparable experimental conditions.

To test this hypothesis, we designed a comprehensive experimental framework involving eleven imputation methods, including five LLM-based approaches, evaluated under MCAR, MAR, and MNAR missingness mechanisms with missing rates of 5\%, 10\%, and 20\%. In addition, we explored the applicability of a foundation model paradigm for missing data imputation. Performance was assessed using the Normalized Root Mean Square Error (NRMSE).

Our results demonstrate that LLMs can be effectively employed for data imputation in tabular data, often achieving superior performance compared to traditional methods on open-source real-world datasets. This advantage is strongly associated with the large-scale background knowledge acquired during their training process. However, this performance gain does not transfer to synthetic or private-data scenarios, where LLMs fail to consistently outperform classical imputation techniques. These findings provide strong evidence that access to knowledge embedded in large-scale internet corpora plays a crucial role in the success of LLM-based imputation, highlighting both their potential and their limitations in data-centric applications.

As a limitation of this work, financial constraints restricted the extent of experimentation with paid LLMs and limited the evaluation across a broader range of LLM-based architectures. Future work could expand the experimental analysis by incorporating additional models to provide a more comprehensive and systematic comparison.

Furthermore, the proposed framework could be extended to different data modalities, enabling a more thorough assessment of the robustness and generalizability of LLM-based data imputation. Another promising direction would be integrating this framework into Automated Machine Learning (AutoML) libraries, potentially improving performance in downstream tasks through more scalable and automated model selection. Finally, an open research avenue involves developing more computationally efficient batch strategies that reduce financial and time constraints while maintaining imputation accuracy.

\section*{Acknowledgements}

This study was financed, in part, by the São Paulo Research Foundation (FAPESP), Brasil. Process Numbers 2021/06870-3 and 2024/23791-8. This work was also financed through national funds by FCT - Fundação para a Ciência e a Tecnologia, I.P., in the framework of the Project UIDB/00326/2025 and UIDP/00326/2025. Additionally, it was supported by the Portuguese Recovery and Resilience Plan (PRR) through project C645008882-00000055-Center for Responsable AI.

\section*{Code availability} \label{subsec:code_ava}

The source code used to implement the proposed methodology and reproduce the experimental results is publicly available at: \url{https://github.com/ArthurMangussi/LLMsImputation}

\bibliographystyle{elsarticle-num}
\bibliography{mybibliography}

\end{document}